%% file: main.tex
\documentclass{article}

\usepackage{arxiv}
\usepackage[utf8]{inputenc}
\usepackage{url}
\usepackage[caption=false, font=normalsize, labelfont=sf, textfont=sf]{subfig}
\usepackage[pdftex]{graphicx}
\usepackage{float}
\usepackage{xcolor}
\usepackage[binary-units=true]{siunitx}

\title{\LARGE \bf
Training Lightweight CNNs for Human-Nanodrone Proximity Interaction from Small Datasets using Background Randomization
}

\author{Marco Ferri, Dario Mantegazza, Elia Cereda, Nicky Zimmerman,\\ Luca M. Gambardella, Daniele Palossi, J\'er\^ome Guzzi, Alessandro Giusti% <-this % stops a space
\thanks{All authors are with the Dalle Molle Institute for Artificial Intelligence (IDSIA), USI-SUPSI, Lugano, Switzerland. DP is also with the Integrated Systems Laboratory (IIS), ETH Z\"urich, Switzerland. NZ is now with the University of Bonn. Contact: \texttt{dario.mantegazza@idsia.ch}. This work is supported by the Swiss National Science Foundation (SNSF) through the NCCR Robotics and by the European Union’s Horizon 2020 Research and Innovation Programs under Grant No. 871743 (1-SWARM).}% <-this % stops a space
}

\begin{document}

\maketitle
\thispagestyle{empty}
\pagestyle{empty}

%%%%%%%%%%%%%%%%%%%%%%%%%%%%%%%%%%%%%%%%%%%%%%%%%%%%%%%%%%%%%%%%%%%%%%%%%%%%%%%%
\begin{abstract}
We consider the task of visually estimating the pose of a human from images acquired by a nearby nano-drone; in this context, we propose a data augmentation approach based on synthetic background substitution to learn a lightweight CNN model from a small real-world training set.  Experimental results on data from two different labs proves that the approach improves generalization to unseen environments.

\end{abstract}
%%%%%%%%%%%%%%%%%%%%%%%%%%%%%%%%%%%%%%%%%%%%%%%%%%%%%%%%%%%%%%%%%%%%%%%%%%%%%%%%

\input{sources/01-Introduction.tex}

\input{sources/03-Implementation.tex}

\input{sources/04-Results.tex}

\input{sources/05-Conclusion.tex}

\bibliographystyle{unsrt}  
\bibliography{biblio}

\end{document}

%% file: sources/01-Introduction.tex
\section{Introduction}\label{sec:introduction}
%What is the problem: we want a CNN that estimates the relative pose of a person; we can train this CNN on data acquired in an optitrack room; if we train only on this data the model does not generalize.

We consider the problem of learning a lightweight model for visually estimating the relative pose of a person from a low-resolution, low-quality image acquired by a nano-drone (i.e., \SI{10}{\centi\meter} diameter and a few tens of grams in weight) flying nearby.  This can be considered a key perception ability for nano-drones interacting with people.
A standard pipeline for solving this problem~\cite{mantegazza2019visionbased, pulp-frontnet} involves: 1) acquiring training sequences in a room equipped with a motion tracking system, while tracking the absolute pose of both the drone and the subject; 2) building a set of training instances, each consisting in a camera frame, and the corresponding ground truth pose of the subject relative to the drone; 3) training a regression model that, given the image, estimates the components of the relative pose.

A critical drawback of this approach is that the training set only contains data acquired in a single environment (or a small set of environments, in case several tracking systems are available).  This does not promote the generalization ability of the resulting models, which will not perform well when used outside the environments used for training.

When training from simulated data, one can solve this problem with \emph{domain randomization} techniques~\cite{tobin2017domain}, which promote generalization by randomizing various aspects of the simulated environment.   Our \textbf{main contribution} (Section \ref{sec:method}) is a similar approach for synthetic augmentation of \emph{real-world} training datasets: in each training frame, we detect people and obtain a binary mask representing pixels belonging to the closest person to the camera, using a pre-trained segmentation model; we use such mask to artificially substitute the background with a random background from a large repository, as an online data augmentation step implemented during training. In Section~\ref{sec:results} we quantify the resulting performance improvements on a testing dataset acquired in a different environment.

% Why it is interesting and important: this is a common pattern in robot perception tasks (one can only collect training data in an instrumented environment, but needs the models to work in a deployment environment which is different). Resource-constrained robots (such as nano-drones) must run very small models and can not afford to run pre-trained large segmentation/detection models such as Mask-R-CNN

%Why do naive solutions fail: One solution to develop models that generalize well is to use domain randomization in simulation and then use sim-to-real techniques: i.e. one generates simulated training data with multiple randomized environments, which builds robustness in the models. In this context (and in general when people and their behaviors are involved) it is not a trivial solution.

%Our main contribution is a sophisticated data augmentation technique, which implements domain randomization from real-world datasets; in particular, ….

\begin{figure}[t]
\centering
\includegraphics[width=0.7\linewidth]{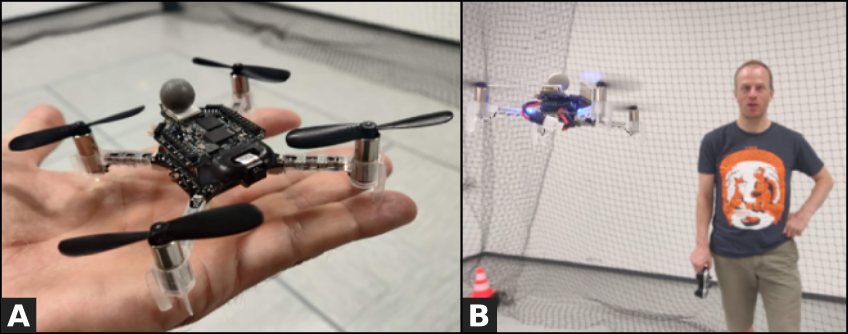}
\caption{The palm-sized robotic platform (A) we envision for the deployment of our work, performing the human-pose estimation task (B).}
\label{tab:prototype}
\end{figure}

\subsection{Related work}\label{sec:related_work}
%The use of deep learning based methodologies for solving the task of autonomous control of unmanned aerial vehicle (UAV) has become the de-facto standard approach for solving such task.
%Nonetheless, deep learning approaches still show major limitations. One of the most important one is the difficulty of these models, mainly neural networks, to generalize well from the training set.

When developing robot perception algorithms, practitioners often train custom deep neural networks from task-specific datasets, which have a limited size; in this context, data augmentation techniques are instrumental to obtain models that generalize well to other environments. Data augmentation artificially generates additional training samples, by applying to the training instances transformations that do not affect the corresponding labels.

%.a common tchnique to improve the generalization ability of neural networks %One of the most common approaches tackling this impediment and improve neural networks' generalization capabilities is data augmentation; image augmentation when applied only on images.  
% When data augmentation is applied to images, thus the data augmentation that is utilised is Image augmentation.
Various approaches have been proposed for augmenting image based datasets~\cite{shorten2019augmentationsurvey}. The most common solutions apply geometric transformations~\cite{xie2020unsupervised} or erase parts of the images 
% to make sure that the model actually looks at the entire image while learning its task
~\cite{wan2013dropconn}. Other methods combine multiple samples to create a new image. In~\cite{Takahashi_2020} and~\cite{summers2019improved}, augmented images are created as a collage of other randomly transformed samples. In~\cite{Lemley_2017}, the authors augment the dataset by creating a combination of images from the same classification category.
% ; the result is similar to the intermediate outputs obtained during the application of a morphing effect \cite{wiki_morphing}
More recently, machine learning approaches have been proposed for automatically learning optimized image augmentation strategies~\cite{zoph2019learning}. 

When collecting training data in simulation, one can additionally exploit Domain Randomization techniques~\cite{tobin2017domain,  tremblay2018training,mehta2019active,loquercio2019deep,yue2019domain,imitation_learning_3d_navigation}, which consists in randomizing various aspects of the simulated environments; this yields models that generalize well to unseen environments, including the real world, since they are not tied to a specific training domain. 

Our approach implements domain randomization to \emph{real-world} datasets, by artificially substituting backgrounds of training frames: the operation does not affect the ground truth label, and can therefore be considered a sophisticated data augmentation step.

%% file: sources/03-Implementation.tex
\section{Background replacement for data augmentation}\label{sec:method}

\subsection{Background replacement}
Our approach relies on a pre-trained Mask R-CNN~\cite{he2018mask} model. Given an image, it returns a bounding box and binary segmentation mask of every object recognized in the image. We apply such model to each training frame, for obtaining a binary mask of the subject that is closest to the camera (which we assume is the one that the drone is interacting with).  More specifically, we obtain all object detections in the frame, then choose the detection with class ``person'' that has the largest bounding box; then, we consider its binary segmentation mask. 
In order to hide the sharp seam between the foreground and background, we can optionally smooth the mask with a Gaussian filter with $\sigma = 1$ px.

%While Mask R-CNN correctly segments different classes of objects in each frame, in order to mask only the correct user, we assume that the tracked one is always the nearest person to the drone's camera. Figure \ref{fig:maskrcnn-dario} shows an example of this.

\begin{figure}[!h]
    \centering
    \includegraphics[width=\linewidth]{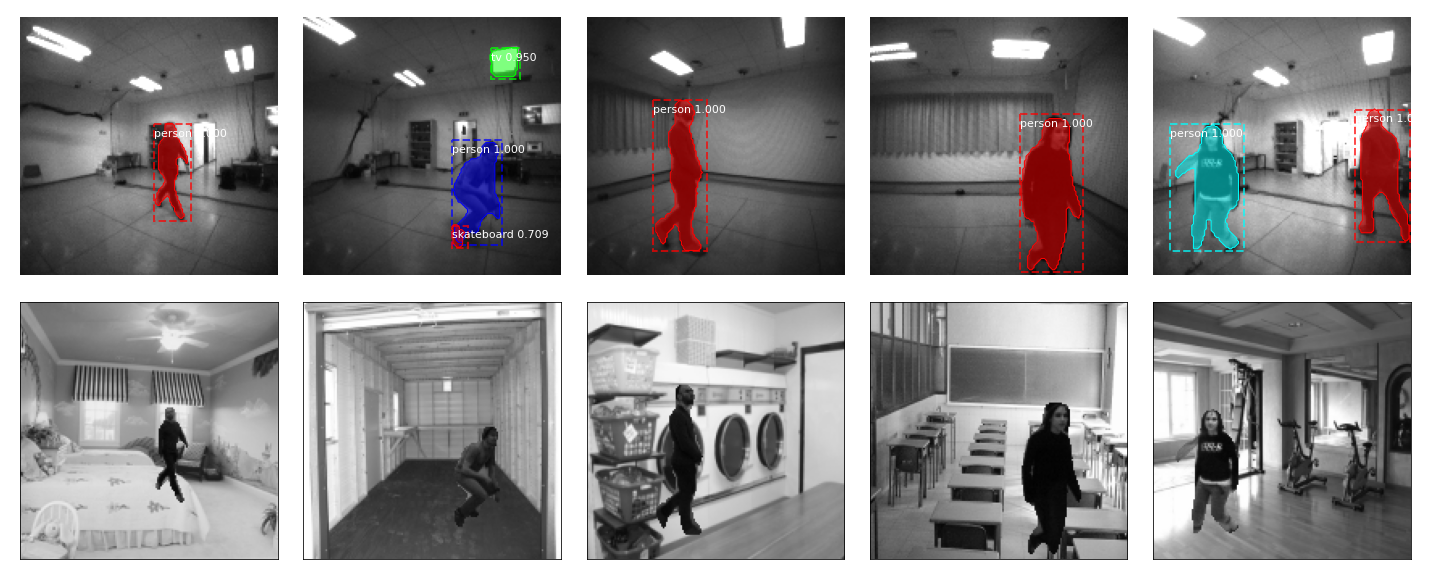}
    \caption{Top row: original training frames with object detections and segmentations (colored masks); note that all training frames are acquired in the same environment. Bottom row: after background randomization.}
    \label{fig:mask-rcnn} 
\end{figure}

The inverse of such mask identifies background pixels and can be used for randomizing the background: for this purpose, we rely on a dataset for Indoor Scene Recognition~\cite{cvpr09}, which contains a total of 15'620 images divided into 67 indoor categories.  After resizing and cropping a random background to the frame size, we paste it over the original frame with the Porter-Duff \emph{over} alpha compositing operator~\cite{porter1984compositing}.  Figure~\ref{fig:mask-rcnn} shows some original training frames with detections overlaid on top, and the frame obtained by randomizing the background of each.

\subsection{Data augmentation pipeline}
\label{subsec:aug-pipeline}
Background masks are computed and stored only once, before training, for each frame in the training set.  This results in a set of training frames ($160\times160$ px, grayscale) and the corresponding background masks ($160\times160$ px, grayscale).

\begin{figure*}[!t]
	\centering
	\includegraphics[width=\linewidth]{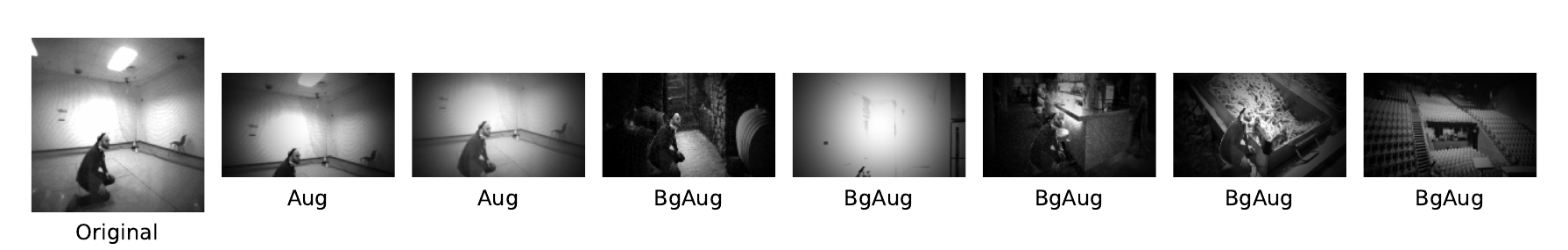}
	\caption[]{The original image (left) and some of its random augmentations. Two (Aug) result from only pitch and photometric augmentation steps. The others (BgAug) result from all the three augmentation steps (Section \ref{subsec:aug-pipeline}).}
	\label{fig:aug-example}
\end{figure*}

During training, every time a frame is selected to be included in a batch, we apply an augmentation function with three sequential steps (Fig. \ref{fig:aug-example}). 
\begin{itemize}
    \item \textbf{Background randomization}, as described above.
    \item \textbf{Pitch augmentation}: we crop a random set of 96 contiguous image rows, which yields a $160 \times 96$ image.  This simulates variations in the pitch of the drone: for example, the top 96 rows of a frame, compared to its middle 96 rows, depict the same scene with an approximate pitch difference of \ang{+14}.
    \item \textbf{Photometric augmentation}: a standard set of randomized photometric augmentations (exposure, gamma correction, dynamic range reduction, blur, additive noise), %as implemented by the Albumentations~\cite{2018arXiv180906839B} python library
    followed by a vignetting effect~\cite{aggarwal2001cosine} with random intensity.  Vignetting significantly affects images acquired by the miniaturized Himax camera implemented in our reference platform.
\end{itemize} 

%% file: sources/04-Results.tex
\section{Experiments}\label{sec:results}

\subsection{Platform}

% ----------- NANO-DRONE STUFF ----------- 
When autonomous navigation capabilities come to the nano-size class of UAVs, there are three main categories of solutions: \textit{i}) offloading the computation to some external power-unconstrained base-station~\cite{kang2019generalization}; \textit{ii}) reducing the onboard workload's complexity to minimal functionalities~\cite{lowlevelcontrol2019}; \textit{iii}) extending the onboard brain either with general-purpose visual navigation engines based on the ultra-low-power (ULP) heterogeneous model~\cite{pulp-frontnet,pulp-dronet1} or with application-specific processors~\cite{navion2019}.

We envision deploying our improved model on a palm-sized UAV that leverages the third solution: the Bitcraze Crazyflie 2.1\footnote{https://www.bitcraze.io/products/crazyflie-2-1} nano-quadrotor, which is open-source but yet compute/memory-limited robotic platform.
To increase the onboard computational energy efficiency, we will leverage the same configuration presented in~\cite{pulp-frontnet}, which extends the basic nano-robot with a pluggable printed circuit board (PCB) called AI-deck.
This PCB plays a crucial role in extending the drone's perception capability, with an ULP (i.e., $\sim\SI{4}{\milli\watt}$) monochrome Himax HM01B0 camera, able to deliver up to \SI{60}{frame/\second} QVGA images.
On the other hand, the AI-deck enables an unprecedented level of computational power for a nano-drone, featuring a general-purpose parallel ULP (PULP) octa-core System-on-Chip (SoC) achieving about \SI{2}{\giga Op/\second}.
Lastly, this companion board offers additional off-chip memory, as much as \SI{8}{\mega\byte} DRAM and \SI{64}{\mega\byte} Flash.

\subsection{Datasets and metrics}

We consider two different environments (EnvA, EnvB), both equipped with a 12-camera OptiTrack motion tracking system.  Both environments are indoor labs with a similar size (approximately $10\times10$ meters) but have significantly different physical characteristics (different walls, furniture, floor, ceiling colors, lighting, etc.).

In each environment, we record a number of sequences featuring a different subject moving in the scene. We manually move around the drone, keeping the subject inside the field of view of camera while capturing different backgrounds: in EnvA, the drone rests flat at a fixed altitude on a cart that we push around, while in EnvB we move the handheld drone freely in space, constantly changing its pitch and roll.

The raw, gray-scale, $160\times160$ images acquired by the drone camera at \SI{60}{frame/\second} are streamed to a PC via WiFi and stored %in a rosbag file, 
together with the absolute poses of drone and subject captured, at the same frequency, by the motion tracking system.
After recording, for each frame we compute the two-dimensional relative pose $(x, y, \phi)$ of the subject with respect to the drone; we form three disjoint datasets: D1 and D2 acquired in EnvA with a different set of subjects, and D3 acquired in EnvB. In total, we acquire more than 12k samples (D1: 2629, D2: 1119, D3: 8737).

% We consider three disjoint datasets: D1 and D2 are acquired in EnvA; D3 is acquired in EnvB.  Each dataset is composed by a number of sequences. Each sequence comprises full frames (grayscale, 160x160 pixels) acquired at 60 FPS, and the corresponding absolute poses of the drone and subject; each sequence features a different subject. 

\subsection{Models and training}

In the following, we use data in D1 to train two CNN regressors, which only differ by the augmentation used during training:
\begin{itemize}
    \item a baseline approach (\textbf{Aug}) which only uses pitch and photometric augmentation;
    \item the proposed approach (\textbf{BgAug}) which uses background randomization, pitch and photometric augmentation.
\end{itemize}

The two models share the same lightweight architecture, which is denoted as $160\times32$ in~\cite{pulp-frontnet}. This architecture has $303 \cdot 10^3$ parameters and can be used for inference onboard the target nano-drone platform at \SI{48}{frame/\second}; it maps a single grayscale $160\times96$ image to the three output pose variables $(x, y, \phi)$.  Note that this architecture is one order of magnitude less compute and memory intensive than the Mask R-CNN model used for augmentation.

The two models are trained for 100 epochs (no early stopping) on a dedicated workstation, equipped with Nvidia GeForce RTX 2080 Ti GPUs.  Each epoch consists of %329
320
steps; each batch contains 64 instances randomly sampled from D1 and augmented using either the Aug or BgAug strategy; on average, during one epoch the model sees 10 randomly augmented versions of each of the 2629 training frames.  20\% of the training set is kept for validation, going through the same augmentation strategy we use for training.

We evaluate each model on both D2 and D3.  For a given frame, we crop the middle 96 rows to get a $160\times96$ image, which is fed to the model; because D2 is shot in the same environment as the training set (EnvA), we expect higher performance than in D3; performance in D3 tests the ability of the model to generalize to a different environment, and its robustness to varying pitch and roll of the camera.

For each output variable ($x$, $y$, $\phi$) we report a standard regression metric, the coefficient of determination $R^2$.  This value indicates the fraction of the variance in the target variable that is explained by the model; it reaches $1.0$ for an ideal regressor, whereas it has a value of $0.0$ for regressors that always return the average of the target variable on the test set.

\subsection{Results and discussion}

\begin{figure*}[!h]
	\centering
	\includegraphics[width=\textwidth]{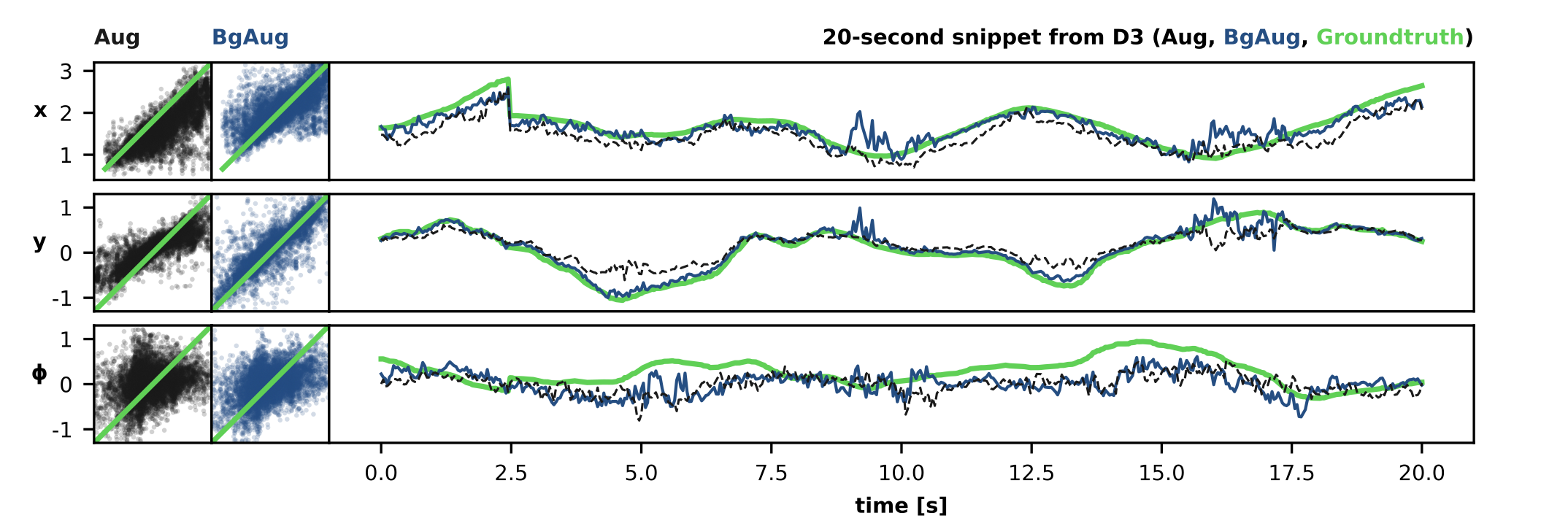}
	\caption[]{For each output variable (rows), we report the ground truth and predictions for Aug (black) and BgAug (blue) models.  Scatter plots on the left compare ground truth (x axis) to predictions (y axis) of the two models on all frames in D3.  Line plots on the right show a time series of the ground truth (thick green) and the two predictions (Aug: dashed black thin line; BgAug: blue thin line) for a short segment in D3.}
	\label{fig:scatter}
\end{figure*} 

Figure~\ref{fig:scatter} compares, for each variable (rows), the relationship between the ground truth and prediction of each of the two models, on all frames of D3. Furthermore, it also shows the evolution of these variables for a short segment of D3.  We observe that, qualitatively, predictions from BgAug match the ground truth more closely than predictions from the Aug model.

\begin{figure}[!t]
	\centering
	\includegraphics[width=0.7\linewidth]{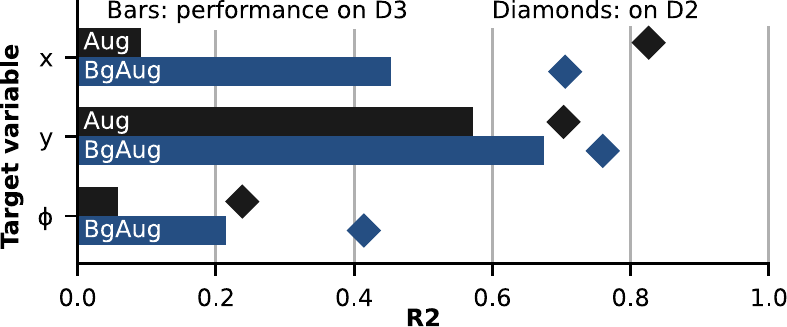}
	\caption[]{$R^2$ on D3 (bars) and D2 (diamonds) for each output variable (y axis) for Aug (black) and BgAug (blue) models.}
	\label{fig:r2-results}
\end{figure}

This observation is quantified in Figure~\ref{fig:r2-results}, which reports the $R^2$ metric for each variable of each model, for both D2 and D3.  As expected, in all cases performance in D2 (diamonds) is better than in D3 (bars), because it more closely resembles the training data.  On D3, BgAug outperforms Aug for all variables, which confirms the effectiveness of the proposed approach.

On D2, background augmentation reduces estimation performance for the $x$ output (i.e. the distance of the subject), whereas it improves the quality of the other two outputs.  We can explain the negative impact on $x$ estimation by observing that estimating the subject distance is closely linked to the pitch of the camera.  When the camera pitches down, the subject head appears higher in the image, and thus the subject appears closer.  We can therefore expect that a model that estimates $x$ well, indirectly also learns to also estimate the camera pitch.  In a given environment, this is done by considering the appearance of the background: background augmentation hinders the model ability to learn this relationship, and thus hinders the estimation  performance of the $x$ variable.

%% file: sources/05-Conclusion.tex
\section{Conclusions}\label{sec:conclusion}
We presented a data augmentation approach based on background randomization for training visual human pose estimation models aimed at resource-constrained nano-drones.  Experimental results on data from two different labs shows that the pipeline yields better generalization ability than a baseline without background randomization.

%% file: main.bbl
\begin{thebibliography}{10}

\bibitem{mantegazza2019visionbased}
Dario Mantegazza, Jérôme Guzzi, Luca~Maria Gambardella, and Alessandro
  Giusti.
\newblock Vision-based control of a quadrotor in user proximity: Mediated vs
  end-to-end learning approaches.
\newblock {\em 2019 IEEE International Conference on Robotics and Automation
  (ICRA)}, May 2019.

\bibitem{pulp-frontnet}
Daniele Palossi, Nicky Zimmerman, Alessio Burrello, Francesco Conti, Hanna
  M{\"u}ller, Luca~Maria Gambardella, Luca Benini, Alessandro Giusti, and
  J{\'e}r{\^o}me Guzzi.
\newblock Fully onboard {AI}-powered human-drone pose estimation on ultra-low
  power autonomous flying nano-{UAVs}.
\newblock {\em arXiv preprint arXiv:2103.10873}, 2021.

\bibitem{tobin2017domain}
Josh Tobin, Rachel Fong, Alex Ray, Jonas Schneider, Wojciech Zaremba, and
  Pieter Abbeel.
\newblock Domain randomization for transferring deep neural networks from
  simulation to the real world, 2017.

\bibitem{shorten2019augmentationsurvey}
Connor Shorten and Taghi Khoshgoftaar.
\newblock A survey on image data augmentation for deep learning.
\newblock {\em Journal of Big Data}, 6, 07 2019.

\bibitem{xie2020unsupervised}
Qizhe Xie, Zihang Dai, Eduard Hovy, Minh-Thang Luong, and Quoc~V. Le.
\newblock Unsupervised data augmentation for consistency training, 2020.

\bibitem{wan2013dropconn}
Li~Wan, Matthew Zeiler, Sixin Zhang, Yann~Le Cun, and Rob Fergus.
\newblock Regularization of neural networks using {DropConnect}.
\newblock In Sanjoy Dasgupta and David McAllester, editors, {\em Proceedings of
  the 30th International Conference on Machine Learning}, volume~28 of {\em
  Proceedings of Machine Learning Research}, pages 1058--1066, Atlanta,
  Georgia, USA, 17--19 Jun 2013. PMLR.

\bibitem{Takahashi_2020}
Ryo Takahashi, Takashi Matsubara, and Kuniaki Uehara.
\newblock Data augmentation using random image cropping and patching for deep
  {CNNs}.
\newblock {\em IEEE Transactions on Circuits and Systems for Video Technology},
  30(9):2917–2931, Sep 2020.

\bibitem{summers2019improved}
Cecilia Summers and Michael~J. Dinneen.
\newblock Improved mixed-example data augmentation, 2019.

\bibitem{Lemley_2017}
Joseph Lemley, Shabab Bazrafkan, and Peter Corcoran.
\newblock Smart augmentation learning an optimal data augmentation strategy.
\newblock {\em IEEE Access}, 5:5858–5869, 2017.

\bibitem{zoph2019learning}
Barret Zoph, Ekin~D. Cubuk, Golnaz Ghiasi, Tsung-Yi Lin, Jonathon Shlens, and
  Quoc~V. Le.
\newblock Learning data augmentation strategies for object detection, 2019.

\bibitem{tremblay2018training}
Jonathan Tremblay, Aayush Prakash, David Acuna, Mark Brophy, Varun Jampani, Cem
  Anil, Thang To, Eric Cameracci, Shaad Boochoon, and Stan Birchfield.
\newblock Training deep networks with synthetic data: Bridging the reality gap
  by domain randomization.
\newblock In {\em Proceedings of the IEEE Conference on Computer Vision and
  Pattern Recognition Workshops}, pages 969--977, 2018.

\bibitem{mehta2019active}
Bhairav Mehta, Manfred Diaz, Florian Golemo, Christopher~J. Pal, and Liam
  Paull.
\newblock Active domain randomization, 2019.

\bibitem{loquercio2019deep}
Antonio Loquercio, Elia Kaufmann, Ren{\'e} Ranftl, Alexey Dosovitskiy, Vladlen
  Koltun, and Davide Scaramuzza.
\newblock Deep drone racing: From simulation to reality with domain
  randomization.
\newblock {\em IEEE Transactions on Robotics}, 36(1):1--14, 2019.

\bibitem{yue2019domain}
Xiangyu Yue, Yang Zhang, Sicheng Zhao, Alberto Sangiovanni-Vincentelli, Kurt
  Keutzer, and Boqing Gong.
\newblock Domain randomization and pyramid consistency: Simulation-to-real
  generalization without accessing target domain data, 2019.

\bibitem{imitation_learning_3d_navigation}
Ahmed Hussein, Eyad Elyan, Mohamed Gaber, and Chrisina Jayne.
\newblock Deep imitation learning for {3D} navigation tasks.
\newblock {\em Neural Computing and Applications}, 04 2018.

\bibitem{he2018mask}
Kaiming He, Georgia Gkioxari, Piotr Dollár, and Ross Girshick.
\newblock Mask {R-CNN}, 2018.

\bibitem{cvpr09}
A.~Quattoni and A.~Torralba.
\newblock Recognizing indoor scenes.
\newblock In {\em 2009 IEEE Conference on Computer Vision and Pattern
  Recognition}, pages 413--420, 2009.

\bibitem{porter1984compositing}
Thomas Porter and Tom Duff.
\newblock Compositing digital images.
\newblock In {\em Proceedings of the 11th annual conference on Computer
  graphics and interactive techniques}, pages 253--259, 1984.

\bibitem{aggarwal2001cosine}
Manoj Aggarwal, Hong Hua, and Narendra Ahuja.
\newblock On cosine-fourth and vignetting effects in real lenses.
\newblock In {\em Proceedings Eighth IEEE International Conference on Computer
  Vision. ICCV 2001}, volume~1, pages 472--479. IEEE, 2001.

\bibitem{kang2019generalization}
K.~{Kang}, S.~{Belkhale}, G.~{Kahn}, P.~{Abbeel}, and S.~{Levine}.
\newblock Generalization through simulation: Integrating simulated and real
  data into deep reinforcement learning for vision-based autonomous flight.
\newblock In {\em 2019 International Conference on Robotics and Automation
  (ICRA)}, pages 6008--6014, 2019.

\bibitem{lowlevelcontrol2019}
Nathan~O Lambert, Daniel~S Drew, Joseph Yaconelli, Sergey Levine, Roberto
  Calandra, and Kristofer~SJ Pister.
\newblock Low-level control of a quadrotor with deep model-based reinforcement
  learning.
\newblock {\em IEEE Robotics and Automation Letters}, 4(4):4224--4230, 2019.

\bibitem{pulp-dronet1}
Daniele Palossi, Antonio Loquercio, Francesco Conti, Eric Flamand, Davide
  Scaramuzza, and Luca Benini.
\newblock A 64-{mW} {DNN}-based visual navigation engine for autonomous
  nano-drones.
\newblock {\em IEEE Internet of Things Journal}, 6(5):8357--8371, 2019.

\bibitem{navion2019}
A.~{Suleiman}, Z.~{Zhang}, L.~{Carlone}, S.~{Karaman}, and V.~{Sze}.
\newblock Navion: A 2-{mW} fully integrated real-time visual-inertial odometry
  accelerator for autonomous navigation of nano drones.
\newblock {\em IEEE Journal of Solid-State Circuits}, 54(4), 2019.

\end{thebibliography}
